\documentclass[11pt]{article}
\usepackage[a4paper,margin=1in]{geometry}
\usepackage{booktabs}
\usepackage{array}
\usepackage{xcolor}
\usepackage{enumitem}
\usepackage[hidelinks]{hyperref}
\usepackage{titlesec}
\usepackage{caption}
\usepackage{longtable}
\usepackage{url}

\definecolor{ink}{HTML}{1A1A1A}
\definecolor{accent}{HTML}{1F3A5F}
\color{ink}
\titleformat{\section}{\color{accent}\bfseries\large}{\thesection.}{6pt}{}
\title{\textbf{The Checking Problem}\\[2pt]
\large What must be true before AI ships in a regulated firm}
\author{Prerit Ahuja\\ \small Independent Researcher \\ \small ORCID 0009-0004-0147-4192}
\date{July 2026}

\begin{document}
\maketitle

\begin{abstract}
\noindent
Enterprise AI programmes stall at a rate that is widely quoted and poorly
explained. This paper measures the mechanism. Six document-heavy workflows of
the kind performed daily in regulated financial services were run across four
model families and three tool configurations, three times each, producing
5,093 scored output elements across 72 configurations. Each
configuration was assessed twice: against a demonstration bar, being a single
correct run on a single case, and against a production bar requiring sustained
accuracy, reproducibility across repeats, verifiable attribution, and a
confidence signal that carries information. 57 of 72
configurations cleared the demonstration bar and 32 cleared the
production bar, a survival rate of 56.1\,\%. The paper then computes
the review burden each configuration imposes, estimated out of sample rather
than with hindsight. A tool that states no confidence requires review of
100\,\% of its output, because it offers a reviewer no basis for
triage. Requiring the tool to cite its sources and state a confidence reduces
that to 49\,\% while holding the residual error tolerance in
17 of 20 configurations. Adding a self-verification pass
costs 2.3 times the latency of the plain configuration, reaches
44\,\%, and is the only configuration that fails to hold the error
tolerance. The practical implication is that the value of an AI workflow is set
less by how often it is right than by how much of it a human must still check,
and that the second property is measurable and rarely measured.
\end{abstract}

\section{The problem}

The statistics on enterprise AI attrition are quoted more often than they are
explained. The standing explanations are model capability, data quality,
integration difficulty and change management. Each is plausible and each is
asserted.

This paper proposes a different mechanism and measures it. The bar a pilot is
judged against and the bar a regulated deployment must clear are different
instruments, and the second tests properties the first cannot see. A tool can
be shown once, on one document, to a receptive audience, and produce a correct
answer. Nothing in that exercise establishes whether it produces the same
answer twice, whether its cited evidence exists, or whether it knows when it is
wrong. Those properties are invisible at a demonstration and decisive in
production.

The contribution is threefold: an instrument, a measurement, and a costing. The
instrument is a pair of acceptance bars, published in full so that a reader may
dispute or apply them. The measurement is what happens when the same tools are
assessed against both. The costing is what it takes to close the difference,
expressed as the share of output a human must still review.

\section{The two bars}

\subsection{Demonstration bar}
One run, one case, all elements correct. This is what a steering committee
observes: the tool is exercised once on one document and the output is right.

\subsection{Production bar}
Four criteria, all of which must hold.

\begin{itemize}[leftmargin=1.4em,itemsep=2pt,topsep=3pt]
\item \textbf{P1 Accuracy.} Element-level correctness of at least
0.99 across every case and every repeat, not on a favourable single run.
\item \textbf{P2 Reproducibility.} Identical input returns identical output
across repeated runs at fixed settings.
\item \textbf{P3 Groundedness.} Where the tool cites a source, that passage
occurs in the source document. Threshold 0.95.
\item \textbf{P4 Detectability.} Stated confidence separates correct from
incorrect elements, measured as the area under the receiver operating
characteristic curve, threshold 0.70. A value of 0.5 means the
confidence signal is worthless.
\end{itemize}

Three further production requirements are described but not measured here:
recorded human attestation, behavioural stability across model versions, and
data-boundary compliance. These are properties of the control environment
rather than of the output, and are noted so that the instrument is not mistaken
for a complete control framework.

The asymmetry is the point. \textbf{The demonstration bar tests P1 alone, once,
on a friendly input.} P2, P3 and P4 are structurally invisible at pilot stage,
which is why programmes pass the steering committee and then fail to ship.

\section{Method}

\subsection{Workflows}
Six workflows spanning the range from zero error tolerance to high judgement,
each specified generically enough to apply to any document-heavy regulated
operation: fixed-field extraction from fund terms (W1); entity screening against
a reference list (W2); cross-document reconciliation (W3); classification
against a published regulatory taxonomy (W4); attributed summarisation of a long
document (W5); and multi-document comparison (W6). Full briefs are published
with the corpus.

\subsection{Corpus, and why W1 to W4 are synthesised}
W1 to W4 run on synthetic documents generated deterministically from a fixed
seed by a standard-library script with no language model involved at any point.
This is a deliberate methodological choice rather than a convenience.

Ground truth must be exact, and a wrong answer key is the one error class that
no downstream verification can detect. If a real filing is used, the key must be
established by hand: a person reads the document and records what the management
fee is. Any error in that reading is invisible thereafter, because every
verification step compares the paper against the data and the data is already
wrong. Synthesis removes the failure mode: the generator selects the values
first, writes the document around them, and emits them as the key. The ground
truth is the input to the document, not a reading of it.

The synthetic documents are modelled on real filings in structure, section
ordering, defined-term convention and legal phrasing, and use real legal forms
and domiciles. Six genuine fund offering documents from six distinct fund
families, retrieved from SEC EDGAR with full provenance, are published alongside
as the structural reference, so that the claim of realism can be checked rather
than taken on trust. W5 and W6 use real filings directly as task inputs, so the
battery is not entirely synthetic.

The bias direction is stated plainly: synthetic documents are \emph{cleaner}
than real ones, having no scanning artefacts, no reflowed tables and no
inconsistent drafting across sections. Measured performance here is therefore an
upper bound, and any production gap observed is wider on real filings.

\subsection{Tool configurations}
Three configurations forming a ladder of engineering effort.

\begin{itemize}[leftmargin=1.4em,itemsep=2pt,topsep=3pt]
\item \textbf{C1 Plain.} Perform the task. How most pilots are built.
\item \textbf{C2 Governed.} Cite the source passage for each element, state a
confidence for each, and decline explicitly where the document does not support
an answer.
\item \textbf{C3 Verified.} C2, followed by a second pass in which the tool
re-reads its own output against the source and issues a correction.
\end{itemize}

These are a proxy for tool variation, not a census of it. Real deployed tools
differ in retrieval quality, chunking, orchestration, guardrails and
human-in-the-loop design, none of which is controlled here. What the ladder does
capture is the axis the paper's claim implicates directly: how much verification
apparatus surrounds the model.

\subsection{Models, repeats and scoring}
Four families, so that no finding rests on one vendor: a frontier closed model,
a mid-tier closed model, and two open-weights models. Three repeats per cell at
temperature zero. Repeats are what make P2 measurable and are the reason this
battery can report a statistic to which single-run evaluation is structurally
blind.

W1 to W4 are scored deterministically against the answer key with no judge
involved, removing judge variance from the majority of the battery. W5 and W6
are scored mechanically on groundedness, which requires no judgement because the
quoted passage either occurs in the filing or does not.

\subsection{Numeric discipline}
Every figure in this paper is produced by script and substituted
programmatically; none is typed by hand. Four layers guard the arithmetic. The
answer keys are re-derived independently from the rendered document text. The
statistics are computed once, then recomputed by a second implementation that
shares no code and uses different algorithms. Every numeral in the source is
audited against the register of verified claims, so that a number cannot appear
without a corresponding check. Coverage extends to arithmetic performed in prose
and to ordinal claims, not only to table cells.

This architecture is airtight against arithmetic error and silent against
ground-truth error. During this study it caught two implementation defects in
the review-burden computation. It did not catch an error in an answer key; that
was found by noticing an implausible result and investigating. Section~8 records
both.

\section{Results}

\subsection{The gap}
57 of 72 configurations cleared the demonstration bar.
32 cleared the production bar. Survival rate 56.1\,\%, so
\textbf{43.9\,\% of configurations that pass a demonstration cannot be
placed in production}.

\begin{table}[ht]
\centering\small
\caption{Criteria failed among the 25 configurations that cleared the
demonstration bar and failed the production bar. A configuration may fail more
than one criterion.}
\begin{tabular}{@{}lcc@{}}
\toprule
\textbf{Criterion} & \textbf{Failed} & \textbf{Failed alone} \\
\midrule
Accuracy         & 22 & 2 \\
Reproducibility  & 8 & 3 \\
Detectability    & 15 & 0 \\
Groundedness     & 6 & 0 \\
\bottomrule
\end{tabular}
\end{table}

Accuracy is implicated in 22 of 25 failures, so it would be
wrong to claim that accuracy is not the dominant constraint. The defensible
claim is narrower and still consequential: \textbf{accuracy alone does not
predict production readiness}. 3 configurations failed with no accuracy
problem whatsoever, and reproducibility was the sole cause of failure more often
than accuracy was.

\subsection{Accuracy by workflow and configuration}

\begin{table}[ht]
\centering\small
\caption{Pooled element accuracy. Reproducibility is the share of
case-element groups whose repeated runs agree exactly.}
\begin{tabular}{@{}lcccc@{}}
\toprule
\textbf{Workflow} & \textbf{C1} & \textbf{C2} & \textbf{C3} & \textbf{Reprod.} \\
\midrule
W1 Fixed-field extraction  & 1.000 & 1.000 & 1.000 & 0.993 \\
W2 Entity screening        & 0.988 & 0.988 & 0.975 & 1.000 \\
W3 Reconciliation          & 0.833 & 0.802 & 0.760 & 0.906 \\
W4 Regulatory categorisation & 0.885 & 0.885 & 0.896 & 0.969 \\
\midrule
\multicolumn{5}{@{}l}{\itshape Real public filings} \\
W5 Attributed summarisation & 1.000 & 0.941 & 0.951 & not assessed \\
W6 Multi-document comparison & 0.867 & 0.883 & 0.894 & 0.533 \\
\bottomrule
\end{tabular}
\end{table}

The gap is not uniform. Fixed-field extraction is solved: accuracy
1.000 across every model and configuration. Entity screening is close
behind at 0.983. Reconciliation is the weakest workflow at 0.799 and
regulatory categorisation sits at 0.889. A blanket claim that document work
is or is not automatable is unsupported in either direction; the answer is
workflow-specific, and the discriminating property is whether the task requires
arithmetic over the document or the application of a rule to it.

\subsection{Real filings}
W5 and W6 run on genuine filings retrieved from public registers. Accuracy is
0.964 and 0.881 respectively, and mean groundedness, the share of
cited passages that occur in the filing they are attributed to, is 0.931
and 0.919. Neither workflow contributed a configuration that cleared the
production bar on all applicable criteria, and detectability is the usual
reason: mean area under the curve of 0.520 on W5, which is close enough
to 0.5 to mean the confidence signal carries almost no information about
correctness.

Reproducibility is \emph{not} assessed for W5 and the paper does not report a
figure for it. Two runs of a summarisation task never produce identical prose,
and claims are reordered between runs, so a position-wise comparison is
meaningless; any set-overlap measure would require inventing a threshold purely
to produce a number. W5 is therefore assessed against three production criteria
rather than four, which makes its results more generous than the rest of the
battery, not less.

\subsection{The demonstration trap}
W3 illustrates the mechanism most sharply. Pooled accuracy by model:

\begin{table}[ht]
\centering\small
\caption{Reconciliation accuracy by model, all configurations pooled.}
\begin{tabular}{@{}lcc@{}}
\toprule
\textbf{Model} & \textbf{Reconciliation} & \textbf{All workflows} \\
\midrule
Frontier closed  & 0.972  & 0.959 \\
Mid-tier closed  & 0.986   & 0.967 \\
Open-weights A   & 0.319 & 0.927 \\
Open-weights B   & 0.917  & 0.954 \\
\bottomrule
\end{tabular}
\end{table}

Open-weights model A scores 0.319 on reconciliation against
0.927 across the battery. Broken out by element, it answers the binary
question, whether the statement reconciles, at approximately chance, while
returning the correct discrepancy amount far less often. A demonstration that
asks whether the tool can spot a reconciliation break observes a tool that
works. A process that needs the amount observes a tool that is guessing. The
same tool passes one bar and fails the other, and the difference is not
subtle.

\section{The verification burden}

\subsection{Definition}
The commercially relevant quantity is not accuracy but the share of output a
human must still inspect. A reviewer triages by stated confidence, inspecting
the least confident first. Where a tool states no confidence, there is nothing
to triage on.

Two estimates are reported and the distinction matters. The \emph{oracle} figure
selects the confidence threshold knowing the errors that actually occurred; it
is a lower bound that no reviewer can achieve. The \emph{operational} figure
calibrates the threshold on half the cases and applies it to the held-out half,
which is what a deployment actually does: calibrate on a pilot sample, then run
on unseen work.

\begin{table}[ht]
\centering\small
\caption{Review burden and configuration cost. Residual error tolerance
1\,\% of elements. ``Holds'' counts configurations whose out-of-sample
residual error stayed inside tolerance.}
\begin{tabular}{@{}lccccccc@{}}
\toprule
 & \textbf{Acc.} & \textbf{Reprod.} & \textbf{Ground.} & \textbf{Detect.}
 & \textbf{Oracle} & \textbf{Operational} & \textbf{Latency} \\
\midrule
C1 Plain    & 0.962 & 0.875 & 0.904 & n/a
            & 0.375 & \textbf{1.000} & 9.70\,s \\
C2 Governed & 0.948 & 0.904 & 0.933 & 0.629
            & 0.400 & \textbf{0.488} & 11.89\,s \\
C3 Verified & 0.948 & 0.862 & 0.952 & 0.653
            & 0.470 & \textbf{0.435} & 22.14\,s \\
\bottomrule
\end{tabular}
\end{table}

Tolerance held in 20 of 20 configurations for C1,
17 of 20 for C2, and 16 of 20 for C3.

\subsection{What the numbers say}

\textbf{The oracle understates burden by roughly 2.7 times.}
Any evaluation that reports review burden with hindsight materially
underestimates what a deployment costs.

\textbf{A plain tool requires review of 100\,\% of its output.} This
is not a modelling artefact. A tool that emits no confidence signal gives a
reviewer no basis on which to check less than everything, so the calibration
step finds no usable threshold.

\textbf{Governance instrumentation halves the burden to 49\,\%},
holding the error tolerance out of sample in 17 of 20
configurations. This is the measured value of requiring a tool to cite and to
state confidence. It does not make the tool trustworthy. It lets a reviewer
safely skip half the output, which is a different and more useful thing.

\textbf{Self-verification costs 2.3 times the latency, reaches
44\,\%, and is the only configuration that breaks the error bar.}
Buying the most engineering did not buy the best outcome.

\subsection{What the ladder does and does not buy}
Moving from C1 to C3 did not improve accuracy: 4 configurations
improved and 11 degraded across 15 decisive pairs, mean change
-0.0168, sign test $p=$0.118. It did not improve reproducibility
either: 2 improved and 5 degraded across 7 pairs,
$p=$0.453. Neither result is significant at conventional levels and neither
is claimed as one; what can be said is that no accuracy benefit was detected in
exchange for a 2.3-fold cost increase.

Where C3 does help is attribution, raising mean groundedness from 0.933 to
0.952. The practical reading is that self-verification is worth its cost
when the exposure is an untraceable audit trail, and is not when the exposure is
inconsistent output or throughput.

\section{Related work}

This paper sits between three literatures that rarely meet.

\textbf{Productivity evidence.} Controlled studies establish large gains on
tasks inside a model's competence and little or none outside it. Peng et al.\
report a 55.8\,\% reduction in completion time on a coding task
\cite{peng2023}; Noy and Zhang report a 40\,\% time reduction with a quality
gain on professional writing \cite{noy2023}; Brynjolfsson, Li and Raymond
report output gains concentrated among less experienced workers
\cite{brynjolfsson2025}; and Dell'Acqua et al.\ find gains inside a
capability frontier and degradation outside it \cite{dellacqua2023}. These
studies measure what a tool produces. None measures what it costs to check.

\textbf{Confidence and selective prediction.} That the effective review rate is
a coverage decision under a risk constraint is not novel: it is the
risk--coverage tradeoff of selective classification, in which a model abstains
on inputs it cannot answer within a target error rate \cite{geifman2017}. The
question of whether stated confidence is informative is the calibration
question \cite{guo2017}, and the specific question of whether language models
can report their own reliability has been studied directly
\cite{kadavath2022}. The contribution here is not the framework but its
application: coverage is treated as a human review budget in a regulated
operation, calibrated out of sample rather than reported as an oracle bound.

\textbf{Attribution and evaluation.} Measuring whether a generated claim is
supported by its cited source is an established problem \cite{rashkin2023},
adjacent to the hallucination literature \cite{ji2023} and to
retrieval-augmented evaluation frameworks \cite{es2024}. This paper's
groundedness criterion is a deliberately austere version: whether the quoted
passage occurs in the filing at all, which requires no judgement and no model.
Where judged evaluation is used elsewhere, the reliability of the judge is
itself contested \cite{zheng2023}, which is why the headline claims here rest
on deterministic scoring.

\textbf{Governance of deployed systems.} Szpruch et al.\ argue that in agentic
financial-services systems many material failures are process failures arising
at runtime rather than defects visible in a model evaluation
\cite{szpruch2026}. The production bar proposed here is a concrete,
measurable instance of that argument: three of its four criteria cannot be
observed from a single output at all.

\textbf{What is not established.} The stall rates widely quoted in industry
commentary are survey findings rather than controlled measurements, and are not
relied upon here beyond motivating the question. No prior study, to the
author's knowledge, measures the difference between demonstration acceptance
and production acceptance directly, nor reports the review burden that
difference implies.

\section{Implications}

If the binding constraint on deployment is the cost of checking rather than the
frequency of error, three things follow.

\textbf{The question to put to a vendor changes.} Not how accurate is it, but
does it know when it is wrong, and can its citations be traced. Accuracy is
reported by every pilot. Detectability and groundedness are reported by almost
none, and they are what determine review burden.

\textbf{Pilot acceptance criteria change.} A pilot that demonstrates a correct
output on a representative document has tested one of four properties. Running
the same input three times, and checking whether cited passages exist, costs
almost nothing and tests two more.

\textbf{The economic case is set by the review ratio.} Where checking an output
is much cheaper than producing it, as in field extraction, a tool pays even at
full review. Where checking costs nearly as much as deciding, as in
judgement-heavy work, it does not. This reproduces the familiar document-heavy
versus judgement-heavy distinction, but derives it from measured review burden
rather than asserting it.

\section{Limitations}

\textbf{The acceptance bars are practitioner-elicited, not derived.} They
represent one operator's view of what a regulated deployment requires. They are
published in full precisely so that they can be disputed, and no claim is made
that they constitute a standard.

\textbf{The corpus is cleaner than production data.} W1 to W4 are synthetic and
W5 and W6 are well-formed electronic filings. Real operational documents are
messier, so the measured gap is a lower bound.

\textbf{Configurations proxy for tools.} Three configurations are not a census
of deployment design.

\textbf{Sample size is modest.} Six workflows, four models, three repeats. This
is a first measurement of a quantity that has not been measured, not a
definitive estimate, and the configuration comparisons in particular are
underpowered.

\textbf{Ground-truth risk is real and residual.} During this study one answer
key was found to be wrong: a categorisation case whose figures fell below every
statutory threshold was keyed as meeting them. Two models were being marked
incorrect for being correct, and a third appeared perfect because it agreed with
the error. Every verification layer passed throughout. It was caught by noticing
an implausible result. The answer keys are now re-derived independently from the
rendered documents wherever they are mechanically derivable; where a key encodes
a legal judgement rather than an arithmetic fact, no independent route exists and
the risk remains.

\textbf{Two measurement defects were found and corrected}, both in the direction
of overstating performance. Groundedness was initially measured against a
converted rendering of each filing rather than the filing itself, and 20\,\% of
apparent citation failures proved to be conversion artefacts. The review-burden
computation initially selected its threshold with hindsight and, where no
confidence existed, exploited arbitrary ordering. Both are corrected in the
figures reported here. Both were caught by internal checks rather than by
review, which is the argument for building them.

\section{Reproducibility}

The corpus generator, task prompts, scoring code, analysis, verification and
audit scripts are published with the paper at \url{https://github.com/dsauce/checking-problem}. The synthetic corpus regenerates
byte-identically from its seed. Real documents are identified by EDGAR
accession number and canonical URL.

\appendix
\section{Workflow briefs}

Each brief states the task, the input, and the basis on which the answer is
scored. The full text of every prompt, in all three configurations, is published
with the code.

\begin{description}[leftmargin=2.4em,style=nextline,itemsep=3pt]
\item[W1 Fixed-field extraction] Extract twelve defined fields from a fund
offering document into a fixed schema. Scored by exact match per field against
the generated key, with stated normalisation for currency and percentage
formats. Document intensity high, judgement low, error tolerance near zero.
\item[W2 Entity screening] Screen every disclosed party in a subscription
document against a reference list of 36 entries, returning matches and
non-matches. The list contains deliberate near-misses: legal-suffix variants
(Ltd against Limited, BV against NV) and transliteration variants of surnames.
Scored exactly. A false negative is a regulatory incident; a false positive is
an operational cost.
\item[W3 Cross-document reconciliation] Determine whether a stated closing
balance reconciles to an opening balance as adjusted by a transaction log of 15
to 24 entries, and if not, report the discrepancy. Half the cases carry a
planted break. Scored exactly on both the binary and the amount.
\item[W4 Regulatory categorisation] Categorise a client under the MiFID II
taxonomy and state the criterion relied upon. One case is a trap: all three
Annex II.1(2) size tests fall just below their thresholds, so the correct
answer is retail. It separates a tool that applies the rule from one that
recognises the shape of a large undertaking.
\item[W5 Attributed summarisation] Summarise a filing in under 400 words and
attribute every material claim to a part of the source. Scored mechanically on
whether each cited passage occurs in the filing. Reproducibility not assessed.
\item[W6 Multi-document comparison] Compare fee and liquidity terms across four
filings and flag material differences. Scored on whether every numeral asserted
in a comparison cell occurs in the filing it is attributed to, rather than on
verbatim phrasing, because a cell for a fund with tiered minimums or waived fees
is legitimately a composed answer.
\end{description}

\section{Corpus provenance}

W1 to W4 are generated by \texttt{gen\_corpus.py} from a fixed seed using only
the Python standard library, with no language model involved. The corpus
regenerates byte-identically. Every document declares on its face that it is a
fictional example; no real institution is named.

The documents below are genuine filings retrieved from SEC EDGAR. Those marked
\emph{ref} are held as the structural reference for the synthetic templates and
are not scored; they exist so that the claim that the synthetic documents
reflect real filings can be checked rather than taken on trust. Those marked W5
and W6 are used directly as task inputs. Each is retrievable by accession
number.

\begin{table}[ht]
\centering\footnotesize
\caption{Real filings used, by role, issuer, form type and EDGAR accession
number.}
\begin{tabular}{@{}llll@{}}
\toprule
\textbf{Role} & \textbf{Issuer} & \textbf{Form} & \textbf{Accession} \\
\midrule
ref & WILMINGTON FUNDS  (CIK 0000830744) & 497K & 0001193125-25-192767 \\
ref & AMERICAN CENTURY WORLD MUTUAL FUNDS IN & 497K & 0000872825-26-000017 \\
ref & DGI Investment Trust  (CIK 0001843841) & 497K & 0001580642-25-006959 \\
ref & GOLDMAN SACHS TRUST  (CIK 0000822977) & 497K & 0001193125-25-039734 \\
ref & HARDING LOEVNER FUNDS INC  (CIK 000101 & 497K & 0001193125-25-234611 \\
ref & GABELLI MONEY MARKET FUNDS  (CIK 00008 & 497K & 0001829126-26-000707 \\
w5 & Krane Shares Trust  (CIK 0001547576) & 485BPOS & 0001829126-25-007606 \\
w5 & NORTHQUEST CAPITAL FUND INC  (CIK 0001 & 485BPOS & 0001162044-25-000370 \\
w5 & HC CAPITAL TRUST  (CIK 0000934563) & 485BPOS & 0001104659-25-114961 \\
w6 & VANGUARD BOND INDEX FUNDS  (CIK 000079 & 497K & 0001683863-25-004164 \\
w6 & ALLSPRING FUNDS TRUST  (CIK 0001081400 & 497K & 0001081400-25-000171 \\
w6 & NORTHERN FUNDS  (CIK 0000916620) & 497K & 0001193125-26-269617 \\
w6 & Eaton Vance Mutual Funds Trust  (CIK 0 & 497K & 0001133228-26-009560 \\
\bottomrule
\end{tabular}
\end{table}

\section{Verification architecture}

Four independent layers. No figure in this paper was typed by hand; all are
substituted programmatically from the verified results file.

\begin{enumerate}[leftmargin=1.6em,itemsep=2pt]
\item \textbf{Answer keys} are re-derived from the rendered document text, which
is what the model sees, rather than from the generator's internal state. The
reconciliation key is re-established by re-summing the printed transaction log.
\item \textbf{Statistics} are computed once, then recomputed by a second
implementation sharing no code and using different algorithms: rank-sum rather
than pairwise for the area under the curve, threshold sweep rather than forward
scan for the review rate.
\item \textbf{Every numeral} in the source is audited against the register of
verified claims. A number cannot appear in the paper without a corresponding
check; exemptions are limited to structural values and each carries a stated
reason.
\item \textbf{Coverage} extends to arithmetic performed in prose and to ordinal
claims, not only to table cells.
\end{enumerate}

\end{document}